\documentclass[10pt,twocolumn,letterpaper]{article}

\usepackage{cvpr}
\usepackage{times}
\usepackage{epsfig}
\usepackage{graphicx}
\usepackage{amsmath}
\usepackage{amssymb}

\usepackage[breaklinks=true,bookmarks=false]{hyperref}

\cvprfinalcopy 

\def\cvprPaperID{14} 

\ifcvprfinal\fi
\begin{document}

\newcommand{\todo}[1]{{\color{red} [TO DO: #1]}}
\newcommand{\comment}[1]{{(\color{blue} #1)}}
\newcommand{\dataset}{WHOI-Plankton\xspace}
\newcommand{\cnnp}{CNN$_{\textrm{P}}$\xspace}
\newcommand{\cnnft}{CNN$_{\textrm{FT}}$\xspace}


\title{\dataset - A Large Scale Fine Grained Visual Recognition Benchmark Dataset for Plankton Classification}

\author{Eric~C.~Orenstein$^{1}$, Oscar~Beijbom$^{2}$, Emily~E.~Peacock$^{3}$, Heidi~M.~Sosik$^{3}$\\
{\tt\small \{e1orenst, obeijbom\}@ucsd.edu}, {\tt\small \{epeacock, hsosik\}@whoi.edu}\\
\\
\{$^{1}$ Scripps Institution of Oceanography, $^{2}$ Department of Computer Science and Engineering\}, \\
University of California - San Diego. $^{3}$ Biology Department, Woods Hole Oceanographic Institution \\
}

\maketitle

\section{Introduction}
Planktonic organisms are of fundamental importance to marine ecosystems: they form the basis of the food web, provide the link between the atmosphere and the deep ocean, and influence global-scale biogeochemical cycles. Scientists are increasingly using imaging-based technologies to study these creatures in their natural habit. Images from such systems provide an unique opportunity to model and understand plankton ecosystems, but the collected datasets can be enormous. The Imaging FlowCytobot (IFCB) at Woods Hole Oceanographic Institution, for example, is an \emph{in situ} system that has been continuously imaging plankton since 2006. To date, it has generated more than 700 million samples. Manual classification of such a vast image collection is impractical due to the size of the data set. In addition, the annotation task is challenging due to the large space of relevant classes, intra-class variability, and inter-class similarity. Methods for automated classification exist, but the accuracy is often below that of human experts. Here we introduce \dataset: a large scale, fine-grained visual recognition dataset for plankton classification, which comprises over 3.4 million expert-labeled images across 70 classes. The labeled image set is complied from over 8 years of near continuous data collection with the IFCB at the Martha's Vineyard Coastal Observatory (MVCO) \cite{olson_submersible_2007, peacock_parasitic_2014}. We discuss relevant metrics for evaluation of classification performance and provide results for a traditional method based on hand-engineered features and two methods based on convolutional neural networks.


\section{Data description}
The IFCB collects images by automatically drawing 5-ml of seawater every 20 minutes from the environment. The seawater is then pumped through a cytometric system; images are only captured of particles that emit chlorophyll fluorescence when they pass through a focused laser beam. Regions of Interest (ROIs) are automatically extracted and saved. To date, the IFCB has collected more than 700 million ROIs at MVCO. A domain expert makes new manual annotations approximately every two weeks by labeling two separate and randomly selected full hours worth of data from the relevant time frame. Since the data to be labeled is selected randomly, each hour of annotated data is a random sample of the plankton population.

The annotated set presented here comprises 3.4 million ROIs spanning 70 classes. There are 100s to 100,000s of samples per class ranging from 100s to 10,000s of pixels. One class of small, undifferentiated particles, the ``mix'' category, encompasses two million ROIs. The number of samples per class is not arbitrary; it is representative of the aggregated natural variability of the class distributions over time.

\section{Methods}
\subsection{Performance metrics}
A relevant problem in plankton ecology is assessing the temporal changes in taxonomic abundance. In the baselines presented below, all data from before 2014 was treated as training data to simulate this experimental goal. Each hour of labeled data in 2014 is then used as an independent test set. This is a realistic deployment scenario since planktonic class distributions are known to change on very short time scales \cite{haury1978patterns, peacock_parasitic_2014}. It is important to note that due to this high variability, not all the classes in the training set are represented in each hour of data (or, for that matter, over a given year).

Oceanographers care about the ability of a classifier to accurately distinguish all classes. This is because the signal is often the fluctuations in abundance of a relatively rare class. The F$_{1}$ score was chosen to compare the methods since it concisely encapsulates class-by-class accuracy. It is defined as the harmonic mean of the precision and recall of a given class. 

To evaluate the three tested methods, the unweighted F$_{1}$ score was computed for each class for each day in 2014. The score was considered in its unweighted form to negate apparent performance boosts from high abundance classes. Note that all 70 classes were not necessarily present in any given day. It is therefore possible for the classifier to assign a label that is not in the label set for that day.

\subsection{Baseline classifiers}
Three classification methods have been applied to this data; a Random Forest (RF) trained on hand-selected features\cite{sosik_automated_2007}, a Convolutional Neural Network (CNN) trained exclusively on plankton data (\cnnp), and a fine-tuned CNN based on a network trained on ImageNet data (\cnnft). Both CNN based methods were implemented using Caffe run on a NVIDIA K40 GPU \cite{jia2014caffe}. All three algorithms were trained with a randomly selected 20$\%$ of all available ROIs from before 2014 ($\sim$650,000).   

The RF uses 229 features including morphological descriptors, texture metrics, invariant moments, etc. For a full list of features and the extraction methodology please see \cite{sosik_automated_2007}. The RF classifier used in the present experiments had 200 trees. 

The \cnnp is a version of the CIFAR10 network modified to use 64x64 pixel images. The images are re-sized according to the longest axis to preserve the aspect, centered in the frame, and padded with the mean pixel value of 10000 randomly selected ROIs from the training set. Training was done in 20000 iterations with a learning rate of 0.001. The network was validated every 1000 iterations on an independent, randomly selected 10$\%$ of all pre-2014 data. 

Finally, the \cnnft was created by fine-tuning the 16-layer CNN developed by the VGG team for the ILSVRC-2014 competition \cite{Simonyan14c}. Samples from the IFCB were placed in the center of a 226x226 grid to match the input dimensions of the VGG network. Samples with dimensions larger than 226 were re-sized in the same way as for the \cnnp network.
 
\begin{figure}[t]
	\centering
    \includegraphics[width=\linewidth]{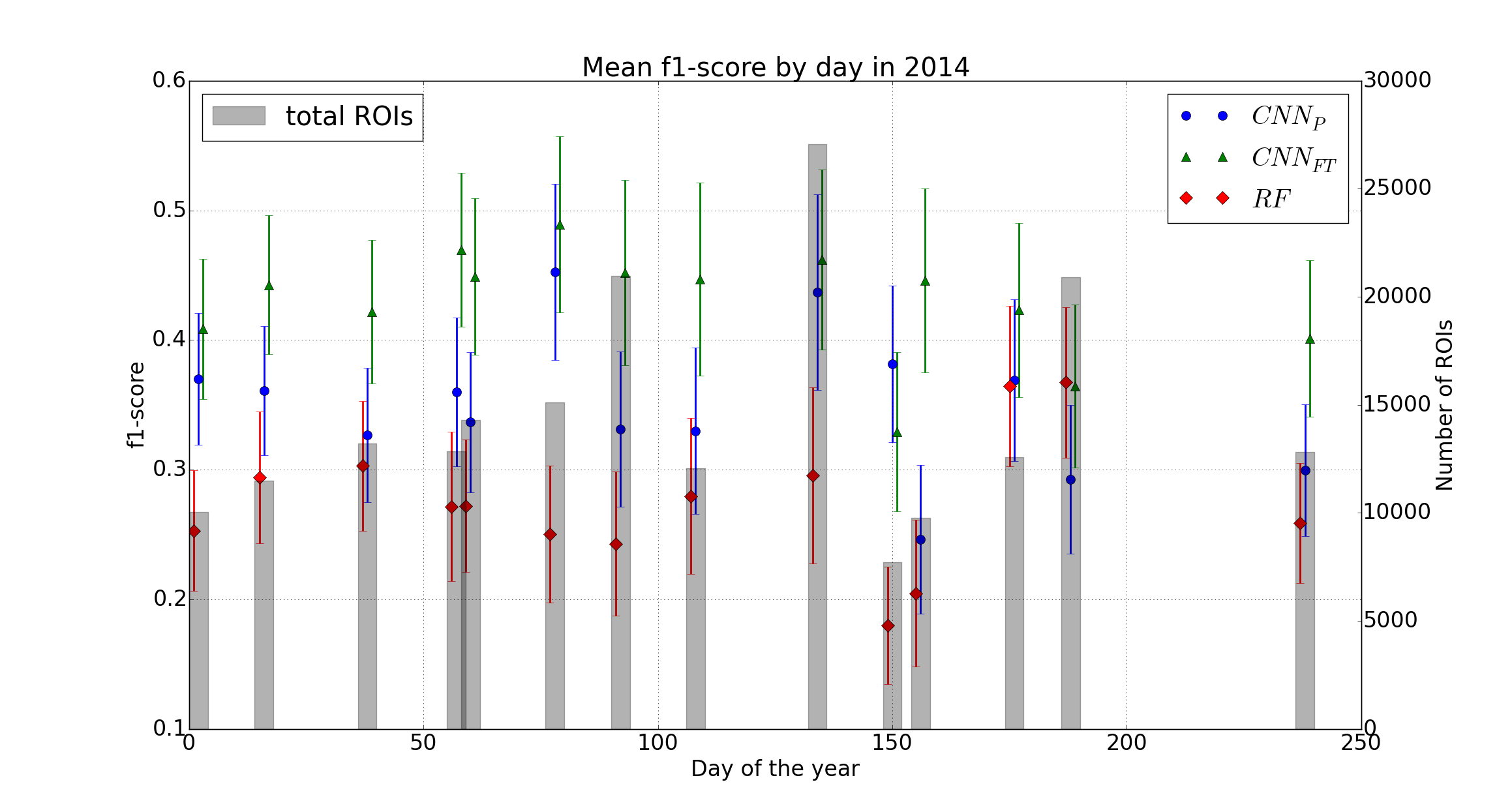}
    \caption{Unweighted, average F$_{1}$ scores for the three methods. The x-axis is day of the year and each point is the score over all labeled samples in that day. Error bars are the standard error all classes each day. The gray histogram bars represent the total ROI counts on each day.}
    \label{f1_all}
\end{figure}

\section{Results}
The accuracy of the three methods were: RF = 90.8\%, \cnnp = 92.8\%, and \cnnft = 93.8\%. The accuracy is generally high because a large majority of the samples belong in the ``mix'' class; a classifier could map all ROIs in the test-set to the ``mix'' class and achieve $\sim$60\% accuracy.

Examining the unweighted, average F$_{1}$ scores for each labeled day in 2014 reveals that the \cnnft outperformed the other two methods on a class-by-class basis (Figure \ref{f1_all}). The RF outperformed the CNN methods on one day. This was because the RF correctly mapped several ROIs to rare classes that the CNNs missed. The mean of average F$_{1}$ scores was taken over all the days in 2014 to aid in comparison. The \cnnft outscored the other methods with 0.42. The \cnnp mean score was 0.36 while the RF got 0.27.

\section{Discussion}
These results provide a starting point for developing classification methods on this unique data set. As suggested by the F$_{1}$ scores, the presented baselines are only somewhat successful in separating out the rare classes. Future work will include  synthetically augmenting training data for rare taxa and developing novel CNN architectures.

{\small
\bibliographystyle{ieee}
\bibliography{cvpr_IFCB}
}

\textbf{Data available at:}

https://github.com/hsosik/WHOI-Plankton

\end{document}